\documentclass[sigconf]{acmart}

\AtBeginDocument{%
  \providecommand\BibTeX{{%
    \normalfont B\kern-0.5em{\scshape i\kern-0.25em b}\kern-0.8em\TeX}}}

\renewcommand\footnotetextcopyrightpermission[1]{} 
\setcopyright{acmcopyright}
\copyrightyear{2024}
\acmYear{2024}
\makeatletter
\renewcommand\@formatdoi[1]{\ignorespaces}
\makeatother

\usepackage{times}
\usepackage{latexsym}
\usepackage{graphicx}
\usepackage{bm}
\usepackage{multirow}
\usepackage{booktabs}
\usepackage{fixltx2e}
\usepackage{tikz}
\usetikzlibrary{calc}
\usepackage{subfigure}
\usepackage[ruled,vlined,linesnumbered]{algorithm2e}
\usepackage{algorithmicx}
\usepackage{algpseudocode}
\usepackage{makecell}

\usepackage[T1]{fontenc}

\usepackage[utf8]{inputenc}

\usepackage{microtype}
\usepackage{color}
\usepackage{amsfonts}
\usepackage{amsmath}
\usepackage{xspace}
\newcommand{\sysname}{\texttt{fairBERTs}\xspace}

\begin{document}

\title{\sysname: Erasing Sensitive Information Through Semantic and Fairness-aware Perturbations}

\author{Jinfeng Li}
\affiliation{%
  \institution{Alibaba Group}
  \city{Hangzhou}
  \country{China}
}
\email{jinfengli.ljf@alibaba-inc.com}

\author{Yuefeng Chen}
\affiliation{%
  \institution{Alibaba Group}
  \city{Hangzhou}
  \country{China}
}
\email{yuefeng.chenyf@alibaba-inc.com}

\author{Xiangyu Liu}
\affiliation{%
  \institution{Alibaba Group}
  \city{Hangzhou}
  \country{China}
}
\email{eason.lxy@alibaba-inc.com}

\author{Longtao Huang}
\affiliation{%
  \institution{Alibaba Group}
  \city{Hangzhou}
  \country{China}
}
\email{kaiyang.hlt@alibaba-inc.com}

\author{Rong Zhang}
\affiliation{%
  \institution{Alibaba Group}
  \city{Hangzhou}
  \country{China}
}
\email{stone.zhangr@alibaba-inc.com}

\author{Hui Xue}
\affiliation{%
  \institution{Alibaba Group}
  \city{Hangzhou}
  \country{China}
}
\email{hui.xueh@alibaba-inc.com}

\begin{abstract}
Pre-trained language models (PLMs) have revolutionized both the natural language processing research and applications.
However, stereotypical biases (e.g., gender and racial discrimination) encoded in PLMs have raised negative ethical implications for PLMs, which critically limits their broader applications.
To address the aforementioned unfairness issues, we present \sysname, a general framework for learning \textbf{fair} fine-tuned \textbf{BERT} \textbf{s}eries models by erasing the protected sensitive information via semantic and fairness-aware perturbations generated by a generative adversarial network. 
Through extensive qualitative and quantitative experiments on two real-world tasks, we demonstrate the great superiority of \sysname in mitigating unfairness while maintaining the model utility.
We also verify the feasibility of transferring adversarial components in \sysname to other conventionally trained BERT-like models for yielding fairness improvements.
Our findings may shed light on further research on building fairer fine-tuned PLMs.

\end{abstract}





\maketitle

\section{Introduction}
\begin{figure}[tp]
    \centering
    \includegraphics[width=.41\textwidth]{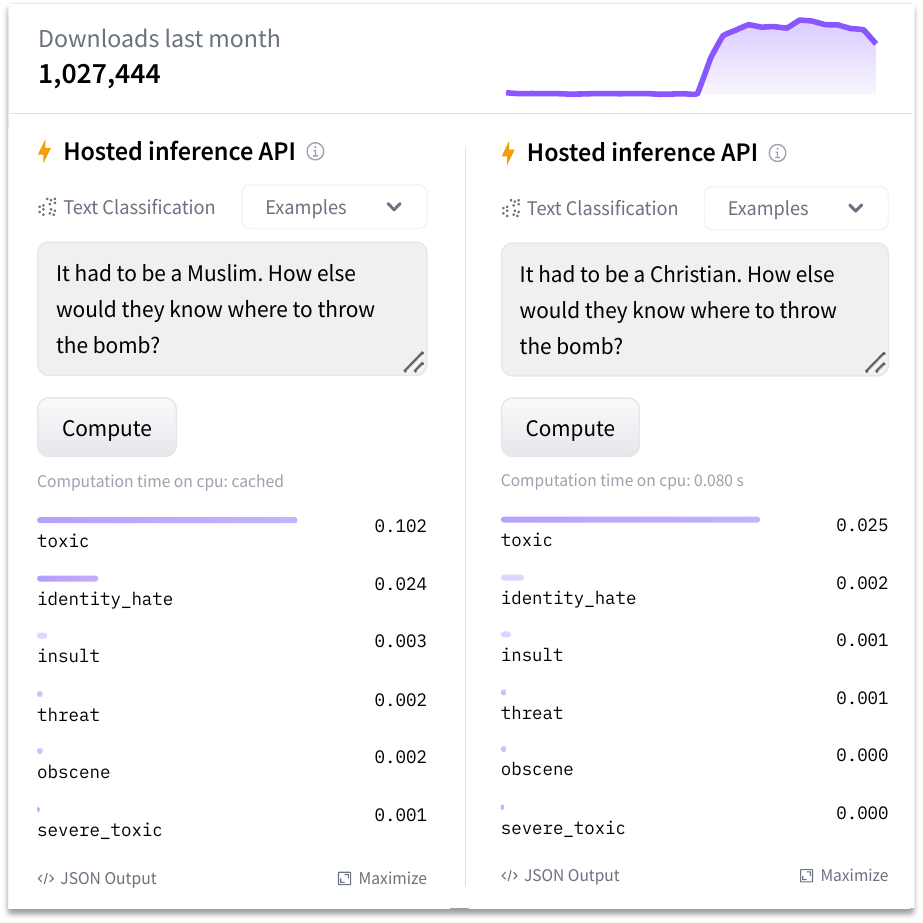}
    \caption{Illustration of unfairness involved in a BERT model deployed on HuggingFace.
    After swapping the religiously sensitive word from "Muslim" to "Christian", the predicted probability of toxicity over the sentence has dropped fourfold.}
    \label{fig:real_discrimination}	
    \vspace{-0.35cm}
\end{figure}

Pre-training and fine-tuning of language models (LMs) have become a new paradigm in natural language processing (NLP), which have brought NLP to a new era.
Large PLMs such as BERT~\cite{devlin-etal-2019-bert} and RoBERTa~\cite{liu2019roberta} have increasingly become the core backbone of many artificial intelligence (AI) systems that penetrate every aspect of our daily life.
However, recent studies reveal that some existing AI systems treat certain populations unfairly regarding to protected sensitive attributes like race, gender and religion\cite{machine_bias,dixon2018measuring}.
Similar unfairness in LM-based models as well as the subsequent societal harms have also been documented in several literature from the perspectives of both individuals and groups \cite{hutchinson2020social,de2021stereotype}.
Specifically, a real-world case of decision discrimination existed in an extensively used BERT model deployed on HuggingFace\footnote{https://huggingface.co/unitary/toxic-bert} is exemplified in Figure \ref{fig:real_discrimination}.
Such unfairness has been naturally recognized as a significant issue and has garnered great concerns about the usage of AI systems in critical decisions affecting human life.

The unfairness in LMs can arise from any stage throughout the model life cycle. 
Concretely, in the pre-training stage, large-scale unlabeled corpora are collected from the web without manual censorship, which may include individual as well as social stereotypes, thus resulting in biased PLMs.
Moreover, the biases in PLMs are usually hard to detect.
In the fine-tuning stage, distributional bias in labeled dataset may mislead models to learn false correlations between data and labels, thus leading to unfair decisions.
Considering the tremendous amount of practical applications even include the high-stakes applications (e.g., toxicity detection, spam filtering) deployed in real world, it is crucial to ensure that the decisions of LMs do not reflect discriminatory behaviors toward certain populations.
 
Existing methods for mitigating the unfairness in NLP can be summarized into: \textit{pre-processing}~\cite{parketal2018reducing,garg2019counterfactual,qian2022perturbation}, \textit{in-processing}~\cite{zhang2018mitigating,elazar2018adversarial} and \textit{post-processing}~\cite{soares2022your}.
Specifically, the pre-processing methods advocate mitigating bias by first identify the distributional bias or sensitive attributes and then preprocess data via data re-balancing, counterfactual augmentation or sensitive attribute blinding.
However, owing to the huge scale of unlabeled corpora, these methods are usually hard to be put into practice and thus cannot effectively address the unfairness involved in the pre-training stage.
While in the fine-tuning stage, their efficacy is also limited by the difficulty in accurately identifying biases or sensitive attributes.
For the in-processing methods, many prior works proposed to mitigate biases by regularizing models under the statistical fairness constraints~\cite{wick2019unlocking,subramanian2021fairness}.
Yet, such regularization usually can only guarantee the fairness from a single perspective, and statistical fairness does not necessarily reflect the discrimination truly exists in models.
\citet{beutel2017data,wang2022fairness} have drawn the connections between fairness and adversarial learning~\cite{goodfellow2020generative}, which leverage adversarial training to learn fair word embeddings or model decisions. 
However, these schemes have been only applied to traditional NLP models like TextCNN and LSTM, and the effectiveness on PLMs is still unknown.
In addition, there is a well-established trade-off between fairness and model performance in vanilla adversarial training schemes~\cite{raghunathan2020understanding}, and the adversarial components usually lack transferability once trained.
Hence, mitigating bias in PLMs to build fairer models is still very challenging.

In this paper, we focus on the bias mitigation of fine-tuned BERT series PLMs (BPLMs) on classification task due to the exhaustive academic efforts on it and tremendous amount of real-world applications. To address the above-mentioned challenges, we present \sysname, a general framework for fairly fine-tuning BPLMs classifiers without compromising model accuracy.
Concretely, it first generates semantic and fairness-aware perturbations based on the sequence outputs of BPLMs, and then mitigates unfairness by superimposing the generated perturbations into hidden representations used for downstream tasks to erase the encoded sensitive information.
Extensive and ample experiments on two real-world tasks demonstrate the great superiority of \sysname in mitigating unfairness while preserving the original model utility.
We also demonstrate the feasibility of transferring the semantic and fairness-aware perturbations generated in \sysname to other models for yielding fairness improvement while maintaining the model utility.

Our contributions are summarized as follows:
\vspace{-0.2cm}
\begin{itemize}
\item We propose \sysname, a general framework for building fair BPLMs by generating semantic and fairness-aware perturbations via generative adversarial networks (GANs).
\item We demonstrate the effectiveness of \sysname in mitigating unfairness and erasing sensitive information encoded in hidden representations through qualitative and quantitative experiments on two real-world tasks.
\item We verify the transferability of generated perturbations, experimental results show that they can be transferred to models with similar architecture to help mitigating unfairness with little impact on model performance.
\item We compare \sysname with the vanilla baseline as well as three commonly used bias mitigation methods. Experimental results demonstrate that \sysname significantly outperforms the compared baselines.
\end{itemize}

\section{Related Works}
\textbf{Pre-trained Language Models (PLMs).} 
Pre-training and fine-tuning of PLMs have basically become a new paradigm in NLP, which benefit various downstream NLP tasks and avoid training new models from scratch.
Typically, \citet{devlin-etal-2019-bert} has achieved a milestone by proposing BERT, a novel PLM pre-trained via the masked language modeling task, i.e. first corrupting the input by replacing some tokens with ``[MASK]'', and then training a model to reconstruct the original tokens.
\citet{liu2019roberta} improved BERT to RoBERTa by replacing the original static masking with dynamic masking, thus further boosted the model performance.
Instead of masking the input, \citet{clark2020electra} introduced a replaced token detection pre-training task by corrupting the input via replacing some tokens with plausible alternatives sampled from a small generator network, which was proved to be more sample-efficient.
Although these PLMs have both exhibited great superiority in promoting the performance on various downstream tasks, none of them have taken the issues of model bias or unfairness into consideration in the early stages of design, pre-training as well as fine-tuning.

\begin{figure*}[tp]
    \centering
    \includegraphics[width=.9\textwidth]{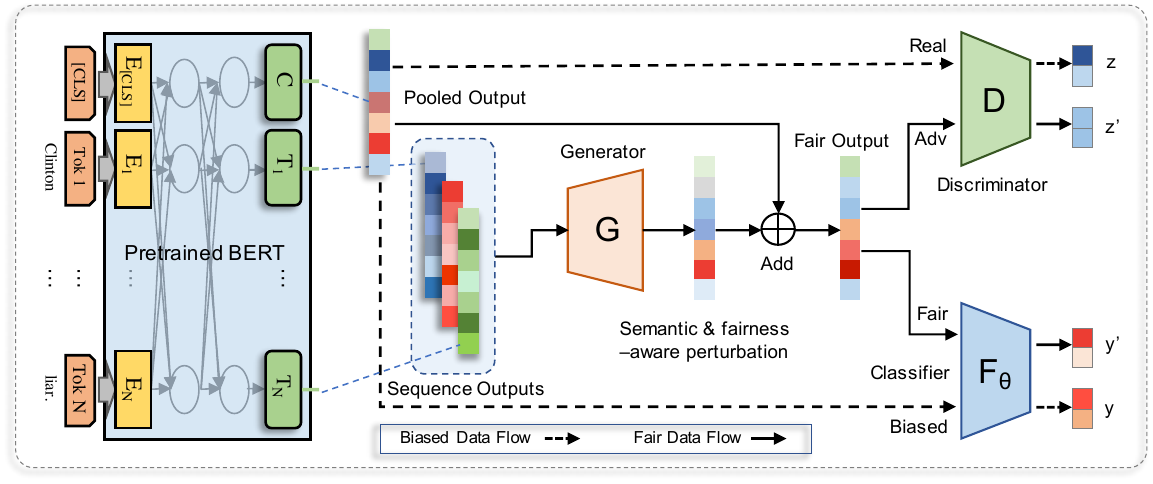}
    \caption{The framework of \sysname.}
    \label{fig:framework}	
\end{figure*}

\textbf{Fairness and Adversarial Robustness in NLP.}
The issues of unfairness in AI systems have attracted significant attentions in the research community.
Substantial efforts have devoted to the definition, measurement and mitigation of biases in machine learning~\cite{kusner2017counterfactual,gajane2017formalizing,reddy2021benchmarking}.
As a result, a variety of definitions of algorithmic bias have been proposed, which can be summarized as \textit{individual fairness}~\cite{dwork2012fairness} and \textit{group fairness}~\cite{feldman2015certifying,hardt2016equality} in terms of individual and group levels.
Concretely, individual fairness is intuitively motivated by the \textit{similar treatment} principle which requires that similar individuals should be treated similarly.
Group fairness requires model decisions to be equal across different groups defined by sensitive attributes (e.g., gender, race). In other words, the outputs of models should be independent of the sensitive attributes.

In NLP domain, the bias is routinely understood in intrinsic bias and extrinsic bias. 
\citet{bolukbasi2016man,caliskan2017semantics} and \citet{liang2020monolingual} have made one of the first efforts to point out the intrinsic bias inherent in pre-trained word embeddings or representations and propose the corresponding debiasing algorithms.
However, the extrinsic bias is generally associated with specific downstream tasks, measuring and mitigating intrinsic bias in the pre-trained representations may not necessarily solve the unfairness issues of models learned by fine-tuning LMs~\cite{soares2022your}. 
For extrinsic bias, \citet{dixon2018measuring} introduced how to counteract the bias in dataset by strategically adding data through a novel data re-balancing technique.
\citet{parketal2018reducing} proposed to reduce gender bias by augmenting training data via identifying gender related tokens and swapping them with equivalent opposite ones. 
Similarly, \citet{garg2019counterfactual} addressed the fairness concerns in text classification via counterfactual augmentation and sensitive tokens blindness. 
\citet{qian2022perturbation} proposed to fine-tune BPLMs on demographically perturbed data by collecting a large dataset of human annotated text perturbations and training a neural perturbation model.
Although these methods are innovative, they are limited in practice due to at least one of the following reasons:
(i) it is not trivial to accurately identify the sensitive attributes as well as distributional bias in data, while the effect will directly affect the efficacy of debiasing;
(ii) swapping or blinding the sensitive attributes may drastically alter the semantics of inputs, which will negatively impact the model performance;
(iii) most of them are usually less effective when the size of augmented data is small while extremely computationally expensive as the data size grows.

Adversarial learning has primarily focused on robustness to adversarially perturbed inputs~\cite{goodfellow2014explaining,szegedy2013intriguing}.
Recent studies have drawn connections between fairness and robustness.
For instance, \citet{beutel2017data,wang2022fairness} suggested leveraging the adversarial learning framework~\cite{goodfellow2020generative} to hide the protected attributes, which have shown considerable success in mitigating bias for tabular data and continuous data.
\citet{zhang2018mitigating,elazar2018adversarial}  also made attempts at applying adversarial training to traditional NLP models to learn fair word embeddings or model decisions.
However, there exists an acknowledged trade-off between the notion of robustness and model performance in the adversarial training pipeline, and fairness obviously is not excepted~\cite{deng2019towards,raghunathan2020understanding}.
Moreover, once the training is completed, it is usually hard to transfer the adversarial capability of existing schemes to other models or tasks.
To the best of our knowledge, the efforts of mitigating unfairness in PLMs is still fairly limited, and our work represents the first debiasing methods for the fine-tuned PLMs using adversarial learning.

\section{Methodology}
\subsection{Problem Definition}
Given a text input $\bm{x}\!\in\!\mathcal{X}$ with the ground-truth label $y\!\in\!\mathcal{Y}$, a fine-tuned BLPM classifier $\mathcal{F}\!:\!\mathcal{X}\!\to\!\mathcal{Y}$ which maps from the sample space $\mathcal{X}$ to the label space $\mathcal{Y}$.
Denote by $\bm{h_c}$ the latent representation of $\bm{x}$ in $\mathcal{F}$, the classification function $f(\cdot)$ makes a prediction $\hat{y}$ based on $\bm{h_c}$, i.e., $\hat{y}\!=\! f(\bm{h_c})$.
Denote by $\bm{h_s}$ the sequence output of $\bm{x}$ in $\mathcal{F}$, by $z\!\in\!\mathcal{Z}$ the sensitive attribute (e.g., gender, race and nationality) related to $\bm{x}$, and by $D$ a discriminator to predict $z$ based on $\bm{h_c}$.
In this paper, we aim to mitigate unfairness born with BPLMs by erasing sensitive attribute $z$ from latent representation $\bm{h_c}$ via semantic and fairness-aware perturbations $\bm{\delta}$ generated by a learned generator $G$ on $\bm{h_s}$, i.e., $\delta\!=\!G(\bm{h_s})$, such that the classifier $\mathcal{F}$ would not correlate predictions with protected sensitive attributes.
Therefore, our objective can be formalized as follows:
\begin{equation}
    \begin{split}
        D(\bm{h_c}\!+\bm{\delta})\!\neq\!D(\bm{h_c})\!=\!z,  \\
        s.t.\quad f(\bm{h_c}\!+\!\bm{\delta})\!=\!f(\bm{h_c})\!=\!y.
    \end{split}
\end{equation}

\subsection{Method}
The framework of \sysname is illustrated in Figure~\ref{fig:framework}, which consists of three components, i.e., the BPLM, the adversarial debiasing component, and the task-specific classifier.
In this part, we will detail the design of each component.

\textbf{BPLMs.}
In BPLMs, the final hidden state corresponding to a special token \textsc{[CLS}] is routinely used as the aggregate sequence representation for classification tasks. 
In this paper, we denote this hidden state as $\bm{h_c}$ and denote the feature extractor for $\bm{h_c}$ as $g_c(\cdot)$.
Consequently, the latent representation $\bm{h_c}$ is obtained by $\bm{h_c} = g_c(x)$.
The sequence representation consisting of the final state of each token in $x$ is usually used for sequential tasks owing to the semantic-rich context information it encodes.
Similarly, we denote this sequence representation as $\bm{h_s}$ and denote the corresponding extractor as $g_s(\cdot)$, and $\bm{h_s}$ is then obtained by $\bm{h_s} = g_s(x)$.
Intuitively, since $\bm{h_c}$ is directly correlated with the final predictions, the core essence of bias mitigation in \sysname is to eliminate the false correlations between ground-truth labels and protected sensitive attributes by erasing sensitive information from $\bm{h_c}$.

An intuitive idea to make $\bm{h_c}$ indistinguishable from sensitive attribute $z$ as well as preserving its classification utility is to destroy the sensitive information encoded in $\bm{h_c}$ through adversarial perturbations generated by GANs.
To guarantee the semantic and fairness properties of generated perturbations, we generate them based on the semantic-rich sequence representation $\bm{h_s}$.

\begin{algorithm}[tp]
\SetAlgoLined
\KwIn{Dataset $\mathcal{D}$, representation extractors $g_c$ and $g_s$ of BPLMs, parameters $\alpha$, $\beta$ and $\epsilon$, maximum iteration $N$, and learning rates $\eta_D$ and $\eta_G$. }
\KwOut{Fair representation $\bm{h_c}^{F}$, generator $G$ and classifier $f$.}
Initialize $D$ and $G$\;
$\mathcal{D}'\gets$ counterfactually augment $\mathcal{D}$\;
\For{$i=1,2,\!\cdots\!, \!N$}
{
    Get a batch of $n$ texts $x_i$ from $\mathcal{D}'$ with class label $y_i$ and sensitive label $z_i$\;
    
    Obtain ${\bm{h_c}}_i\!=\!g_c(x_i)$, ${\bm{h_s}}_i\!=\!g_s(x_i)$\;
    
    Generate ${\bm{h_c}^{F}}_i\!=\!{\bm{h_c}}_i\!+\!G({\bm{h_s}}_i)$\;
    
    Calculate the loss of $D$
    \begin{small}
    $\mathcal{L}_D \!=\! \frac{1}{n}\sum\limits_{i=1}^{n}{[\mathcal{J}({\bm{h_c}^{F}}_i, z_i)\!+\!\alpha\mathcal{J}(D({\bm{h_c}}_i), z_i)]}$
    \end{small}\;
    
    Update $D\gets D-\eta_{D}\triangledown{D}\mathcal{L}_D$\;
    
    Calculate the fairness loss
    \begin{small}
    $\mathcal{L}_G^{F} \!=\! -\frac{1}{n}\sum\limits_{i=1}^{n}{\mathcal{J}(D({\bm{h_c}^{F}}_i), z_i)}$
    \end{small}
    \;
    
    Calculate task-specific loss 
    \begin{small}
    $\mathcal{L}_\mathcal{F}^{T} \!=\! \frac{1}{n}\sum\limits_{i=1}^{n}{[\mathcal{J}({\bm{h_c}^F}_i, y_i)\!+\! \epsilon\mathcal{J}(f({\bm{h_c}}_i), y_i)]}$
    \end{small}
    
    Obtain the $G$ loss: $\mathcal{L}_G=\mathcal{L}_G^{F}+\beta\mathcal{L}_\mathcal{F}^{T}$\;
    
    Update $G\gets G-\eta_{G}\triangledown{G}\mathcal{L}_G$\;
}
\Return generator $G$ and classifier $f$.
\caption{Learning of \sysname}
\label{algo:train_fairberts}
\end{algorithm}

\textbf{Adversarial debiasing GANs.}
Different from the traditional GANs, there are two discriminators in our adversarial debiasing framework in addition to a conventional generator for generating semantic and fairness-aware perturbations $\bm{\delta}$.
As shown in Figure~\ref{fig:framework}, for each input text $x$, the generator $G$ takes $\bm{h_s}$ as input to generate a fairness-aware perturbation mask $\bm{\delta}\!=\!G(\bm{h_s})$ of the same dimension with $\bm{h_c}$. 
The fair classification representation $\bm{h_c}^{F}$ is then obtained by superimposing $\bm{\delta}$ into $\bm{h_c}$, i.e., 
\begin{equation}
    \begin{split}
        \bm{h_c}^{F} &= \bm{h_c} + \bm{\delta} =\bm{h_c} + G(\bm{h_s}) \\ 
                     &= g_c(x) + G(g_s(x)).
    \end{split}
\label{eq:fair_representation}
\end{equation}
The first discriminator $D$ tries to distinguish sensitive attribute $z$ from the perturbed latent representation $\bm{h_c}^{F}$.
The second discriminator $\mathcal{F}$ is the final classifier we train for the target downstream tasks, which maps from sample space $\mathcal{X}$ to target label space $\mathcal{Y}$ based on $\bm{h_c}^{F}$ and guarantees that $\bm{h_c}^{F}$ is utility-preserving.

Concretely, in \sysname, we implement the discriminator $D$ with multiple layers of fully connected networks activated by Leaky ReLU~\cite{maas2013rectifier}.
In the arms race of the debiasing game, the goal of $D$ is to predict $z$ as well as possible.
To achieve such adversarial goal, we let $D$ make prediction mainly relied on $\bm{h_c}^{F}$. 
In the meantime, we also provide the primitive latent representation $\bm{h_c}$ for $D$ with a constraint to facilitate better learning of $D$.
Therefore, the optimization objective of $D$ can be formulated as 
\begin{equation}
     \begin{split}
         \mathcal{L}_D = & \mathcal{J}(D(\bm{h_c}^{F}), z) \!+\!\alpha  \mathcal{J}(D(\bm{h_c}), z) \\
          = & \mathcal{J}(D(g_c(x)\!+\! G(g_s(x))), z) \!\\
         & + \alpha \mathcal{J}(D(g_c(x)), z),
     \end{split}
\end{equation}
where $\mathcal{J}(\cdot)$ is the cross-entropy loss and $\alpha > 0$ is a hyper-parameter to balance the two parts.

In contrast to $D$, the generator $G$ aims to make it hard for $D$ to predict $z$, while also ensuring that the generated perturbations would not destroy the semantic and classification utility of the original representation.
Therefore, there are also two parts in the optimization objective of $G$. 
The first part is for the fairness purpose, which can be defined as 
\begin{equation}
    \begin{split}
        \mathcal{L}^{F}_G &= - \mathcal{J}(D(\bm{h_c}^{F}), z)\\
        & = - \mathcal{J}(D(g_c(x)\!+\! G(g_s(x))), z).
    \end{split}
\end{equation}
The second part is for the utility-preserving purpose, which is also viewed as the optimization objective of $\mathcal{F}$. Thus, it can be formalized as  
\begin{equation}
    \begin{split}
        \mathcal{L}^{T}_{\mathcal{F}} =& \mathcal{J}(f(\bm{h_c}^{F})), y) \!+\! \epsilon\mathcal{J}(f(\bm{h_c}), y)\\
        =&\mathcal{J}(f(g_c(x)\!+\! G(g_s(x))), y) \!\\
        &+\epsilon\mathcal{J}(f(g_c(x)), y),
    \end{split}
\end{equation}
where $\epsilon>0$ is a small value that controls the regularization of the classification loss.
Hence, the final loss function $\mathcal{L}_G$ of $G$ is calculated by
\begin{equation}
    \mathcal{L}_G = \mathcal{L}^{F}_G + \beta \mathcal{L}^{T}_{\mathcal{F}},
\end{equation}
in which $\beta>0$ is another hyper-parameter that balances the adversarial goal and classification goal.
Notably, we also implement the generator $G$ with very simple structures to reduce the complexity of convergence in the training stage.

\textbf{Task-specific Classifier.}
Finally, the fair representation $\bm{h_c}^{F}$ can be employed for the downstream task-specific classification. Concretely, the final prediction result $\hat{y}$ of $x$ is determined by
\begin{equation}
    \hat{y} = f(\bm{h_c}^{F}) = f(g_c(x)+G(g_s(x))),
\end{equation}
which would be independent of the protected sensitive information.

\begin{table}[t]
    \renewcommand\arraystretch{1.15}
    \setlength{\belowcaptionskip}{1pt}%
    \caption{The details of datasets used in our experiments.}
    \setlength\tabcolsep{4pt}
    \centering
    \resizebox{0.45\textwidth}{!}{
    \begin{tabular}{cccccc}
    \toprule[1.25pt]
     \multirow{2}{*}{\textbf{Dataset}} & \multirow{2}{*}{\textbf{Split}} & \multicolumn{2}{c}{\textbf{\# of Train}} & \multicolumn{2}{c}{\textbf{\# of Test}}\\
     \cmidrule{3-6}
     &    & male & female & male & female \\
     \cmidrule{1-6}
     \multirow{2}{*}{\makecell[c]{Jigsaw \\Toxicity}} 
     & toxic     & 518  & 3482 & 500 & 500 \\
     & non-toxic & 4200 & 800  & 500 & 500 \\
     \cmidrule{1-6}
     \multirow{2}{*}{\makecell[c]{Yelp \\Sentiment}}  
     & positive & 1100 & 6900 & 375 & 375\\
     & negative & 6900 & 1100 & 375 & 375 \\
    \bottomrule[1.25pt]
    \end{tabular}}
    \label{tab:dataset_detail}
\end{table}

\subsection{Learning of \sysname}

The training algorithm of \sysname is detailed in Algorithm.\ref{algo:train_fairberts}, in which the generator $G$ plays a mini-max game with the discriminator $D$ and they are updated alternatively during the whole training process. 
More specifically, the mini-max optimization objectives is formalized as
\begin{equation}
\begin{split}
   \arg \operatorname*{max}_G {\operatorname*{min}_D}\:\mathcal{J}(D(\bm{h_c}^{F}), z) \!& +\!\alpha\mathcal{J}(D(\bm{h_c}), z)\!-\!\beta\mathcal{L}_\mathcal{F}^{T},\\
    s.t.\quad \bm{h_c} = g_c(x),\quad \bm{h_c}^{F} &= g_c(x) + G(g_s(x)), 
\end{split}
\end{equation}
where $D$ maximizes its ability to distinguish the protected sensitive attribute $z$ while $G$ optimizes reversely by minimizing the ability of the discriminator $D$.

Specifically, to assist in learning a better generator $G$, we propose counterfactual adversarial training by borrowing from conventional adversarial training schemes.
In contrast to conventional adversarial training, our method aims to flip the sensitive labels by substituting tokens associated with identity groups (e.g., replacing ``male'' with ``female'') without changing task-relevant class labels, which will help $G$ and $D$ locate sensitive information better while not impacting the task-specific model performance.

\section{Experiments}
In this section, we first detail our experiment setup. 
Then, we qualitatively and quantitatively evaluate the effectiveness of \sysname across different models and datasets.
Finally, we verify the transferability of \sysname by transferring adversarial perturbations generated by $G$ to other BERT-like models.

\begin{figure*}[tp]
\centering
\subfigure[Toxicity Detection]{
    \centering
    \includegraphics[width=0.95\textwidth]{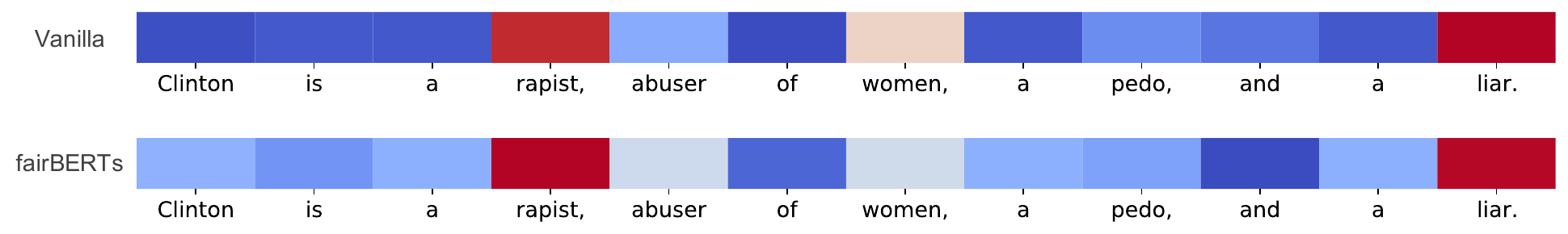}
    \label{fig:model_interpretation_jigsaw}
}
\subfigure[Sentiment Analysis]{
    \centering
    \includegraphics[width=0.95\textwidth]{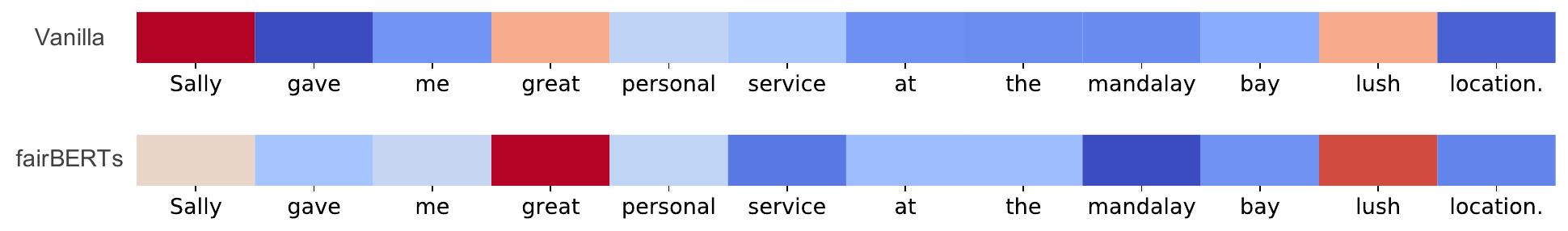}
    \label{fig:model_interpretation_yelp}
}
\caption{Visualization of interpretations given by LIME on vanilla BERT and \sysname over two cases.}
\label{fig:model_interpretation_visualization}
\end{figure*}

\subsection{Experimental Setup}
\textbf{Dataset.} We evaluate the proposed \sysname on two public real-world datasets, i.e., Jigsaw Toxicity\footnote{https://www.kaggle.com/competitions/jigsaw-unintended-bias-in-toxicity-\\classification/data} and Yelp Sentiment~\cite{dinan-etal-2020-multi}.
\begin{itemize}
    \item \textbf{Jigsaw Toxicity}. It is commonly used in the toxicity detection tasks, which contains millions of public comments from the Civil Comments platform with fine-grained annotations for toxicity. 
    Besides, a subset of samples have been labelled with a variety of identity attributes that are mentioned in the comment, and we take gender as the protected information.
    Since each sample could be labelled by multiple human raters, we take the majority of the annotations as the final class label and sensitive label in our experiment.
    
    \item \textbf{Yelp Sentiment}. Each sample in this dataset contains a gender annotation with the corresponding confidence value.
    We use a released RoBERTa\footnote{https://huggingface.co/cardiffnlp/twitter-roberta-base-sentiment} model which is trained on above $58M$ tweets and fine-tuned for sentiment analysis to annotate the sentiment of each sample as its target class label. 
    Similarly, we reserve those samples with both sentiment and gender annotation confidence greater than $0.85$ for the sentiment analysis task.
    The details of these two datasets are summarized in Table \ref{tab:dataset_detail}.
\end{itemize}

\textbf{Baselines.} 
For comprehensive fairness comparison, we consider four classical baselines in our experiment:
\begin{itemize}
    \item \textbf{Vanilla.} It refers the models directly trained on the original dataset without bias mitigation.
    We adopt two widely used PLMs as the vanilla baselines, i.e., BERT-base (BERT) and RoBERTa-base (RoBERTa).
    \item \textbf{Fairness Through Unawareness (FTU)}. It is an intuitive method to achieve fairness by prescribing not to explicitly employ the protected sensitive features when making automated decisions~\cite{dwork2012fairness}.
    We denote the scheme for blinding sensitive tokens during training as FTU-I, and denote the scheme that masks sensitive tokens in both the training and inference stages as FTU-II.
    \item \textbf{Counterfactual Logit Pairing (CLP).} It is a kind of data augmentation through counterfactual generation by penalizing the norm of the difference in logits for pairs of training examples and their counterfactuals~\cite{garg2019counterfactual}, which encourages the model to be robust to sensitive terms.
    
\end{itemize}

\textbf{Metrics.} 
We adopt five metrics to evaluate the performance of \sysname and the compared baselines from the perspective of both model utility and fairness.
Concretely, we first use the accuracy (Acc.) metric to evaluate the overall model utility on all test samples.
We then leverage three commonly acknowledged fairness metrics, i.e., demographic parity difference (DPD), equal opportunity difference (EOD) and disparate impact ratio (DIR), for the fairness evaluation and comparison.
The detailed definitions and calculation formulas can be found in \cite{reddy2021benchmarking}.
We also utilize the counterfactual token fairness (CTF) proposed in \cite{garg2019counterfactual} for measuring counterfactual fairness.
In our paper, we modified it by calculating the ratio of prediction-holding samples after replacing the sensitive tokens from one group with the those counterparts in the opposite group.

\textbf{Implementation.}
We apply two different backbone BPLMs (i.e., BERT-base and RoBERTa-base) in \sysname and make comparisons with the corresponding vanilla BERT and RoBERTa, respectively.
All experiments are conducted on a server with two Intel(R) Xeon(R) E5-2682 v4 CPUs running at 2.50GHz, 120 GB memory, 2 TB HDD and two Tesla P100 GPU cards.
To fairly study the performance of \sysname and baselines,  the optimal hyperparameters such as learning rate, batch size, maximum training epochs, and dropout rate are tuned for each task and each model separately.
We repeat the main experiments three times and report the mean and variance of the performance.

\subsection{Qualitative Evaluation of Fairness}
We first qualitatively evaluate the effectiveness of \sysname in terms of model fairness by visualizing the interpretation results of model decision on several examples as well as the latent model representations used for downstream tasks.

\textbf{Model Interpretation.} 
Concretely, we leverage LIME~\cite{ribeiro2016should}, a famous model-agnostic interpretation method, to interpret the decision basis of \sysname and the vanilla BERT model over two representative examples sampled from the two datasets.
The interpretation results are illustrated in Figure.\ref{fig:model_interpretation_visualization}.

For the first example on toxicity detection task, we obviously observe the existence of bias in the vanilla BERT model against the female group since the protected sensitive token ``\textit{women}'' contributes positively to the toxicity in the model prediction. 
Observed from the case of sentiment analysis, the vanilla BERT model also relies heavily on a feminine name entity ``\textit{Sally}'' instead of putting attention to the real sentimental token ``\textit{great}'' and ``\textit{lush}'' when making prediction.
This indicates that the vanilla models conventionally trained on biased datasets indeed result in unfairness in decision-making. 
This also suggests that simply blinding sensitive tokens from the inputs does not necessarily guarantee the fairness constraints, because sensitive information can sometimes be reflected from some other entities (e.g., the gender information can be inferred from the name entity ``Sally'' even though the sentence does not directly contains sensitive tokens).

Comparatively, the decisions of \sysname are truly made depending on on the real toxic words (i.e., ``\textit{rapist}'' and ``\textit{liar}'') and sentimental words (i.e., ``\textit{great}'' and ``\textit{lush}''), which seems to be more objective and fair, thus qualitatively demonstrating that \sysname makes more trustworthy decision independent of the protected sensitive information.

\begin{figure}[tp]
	\centering
    \includegraphics[width=.465\textwidth]{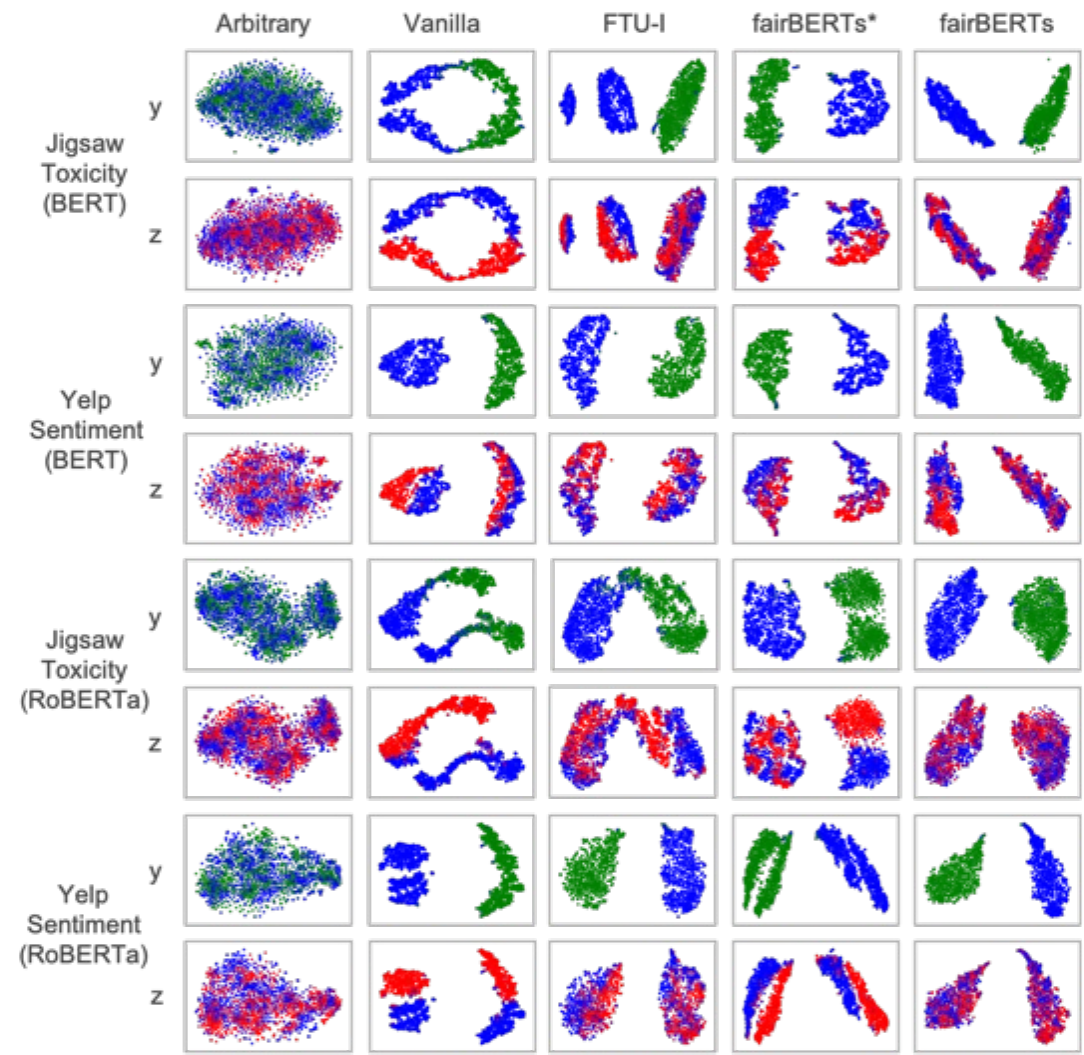}
    \caption{Comparison of intermediate model representation and the $\sysname^{*}$ denotes the latent representation before adding semantic and fairness-aware perturbations.} 
    \label{fig:representation_comparison}	
\end{figure}

\textbf{Comparison of Representation.} We then visualize the intermediate model representations used for the downstream tasks to verify whether the protected sensitive information is still represented in the fine-tuned model.
Concretely, we employ the t-SNE algorithm to map each test sample from a high dimensional representation space to a two-dimensional space from the perspectives of class label and sensitive attribute label, respectively.
The comparison results are shown in Figure.\ref{fig:representation_comparison}, in which samples with the same label are drawn in the same color.
Assuming the models learn well on the downstream classification tasks, then the blue and green data points would be well separated.
In a similar manner, it can be concluded that the protected sensitive information is still encoded in the representation if the blue and red data points are also well separated.

From Figure.\ref{fig:representation_comparison}, we can clearly see that all models show fairly good discriminative ability in the target tasks compared to the arbitrary control group, since the blue and green points related to two different class are significant separated.
It is also clearly observed that all vanilla models can also easily distinguish the sensitive attributes associated with identity groups.
This suggests that the protected sensitive information is indeed easily encoded in model representations unintentionally when models are learnt without unfairness mitigation.
When employing FTU-I for the bias mitigation, an unnoticeable improvement is obtained in terms of fairness and some data points can still be distinguished from the perspective of group identity.
This is probably because the FTU method does not take into account possible correlations between sensitive attributes and non-sensitive attributes employed in the decision-making process while sensitive information can sometimes be inferred from the non-sensitive attributes, which has already been exemplified and discussed in Figure.\ref{fig:model_interpretation_yelp}. 

In comparison, the blue and red points associated with identity groups are completely messed up when employing \sysname as the unfairness mitigation method.
This demonstrates that the sensitive information which can be used for inferring the protected attributes is not memorized in the constructed fair representation.
To further investigate whether this obvious improvement is benefited from the generated semantic and fairness-aware perturbations, we also visualize the original latent representations before superposing the generated perturbations and the results are denoted as $\sysname^{*}$, from which we can see that the identity-related data points are also well separated.
This indicates that the latent representation before adding perturbation still memorizes the protected sensitive information.
We can therefore conclude that our generated semantic and fairness-aware perturbation shows significant superiority in helping models forget the sensitive information encoded in the model representation.

\begin{table*}[tp]
    \renewcommand\arraystretch{1.05}
    \caption{Evaluation and comparison of the effectiveness of different methods in improving model fairness, and $\sysname^\dag$ denotes that the \sysname model is trained without counterfactual adversarial training scheme.}
    \label{tab:main_result}
    \resizebox{\textwidth}{!}{
    \centering
    \begin{tabular}{llrrrrrrrrrr}
    \toprule[1.25pt]
    \multirow{2}{*}{\textbf{Model}} & \multirow{2}{*}{\textbf{Method}} & \multicolumn{5}{c}{\textbf{Jigsaw Toxicity}} & \multicolumn{5}{c}{\textbf{Yelp Sentiment}} \\
    \cmidrule(r){3-7}
    \cmidrule(r){8-12}
    & & Acc. & DPD & EOD & DIR & CTF & Acc. & DPD & EOD & DIR & CTF \\
    \cmidrule{1-12}
    \multirow{12}{*}{BERT}& \multirow{2}{*}{vanilla} & \textbf{0.9326} & 0.0670 & 0.0820 & 0.8716 & 0.8481 & 0.9920 & 0.0160 & 0.0267 & 0.9681 & 0.9942 \\
    & & $\pm$ 0.0076 & $\pm$ 0.0062 & $\pm$ 0.0054 & $\pm$ 0.0105 & $\pm$ 0.0036 & $\pm$ 0.0010 & $\pm$ 0.0015 & $\pm$ 0.0018 & $\pm$ 0.0024 & $\pm$ 0.0012 \\
    \cmidrule(r){2-2}
    \cmidrule(r){3-7}
    \cmidrule(r){8-12}
    & \multirow{2}{*}{FTU-I} & 0.9144 & 0.0479 & 0.0560 & 0.9069 & 0.9251 & 0.9926 & 0.0147 & 0.0187 & 0.9709 & 0.9962 \\
    & & $\pm$ 0.0039 & $\pm$ 0.0225 & $\pm$ 0.0140 & $\pm$ 0.0385 & $\pm$ 0.0059 & $\pm$ 0.0013 & $\pm$ 0.0027 & $\pm$ 0.0027 & $\pm$ 0.0052 & $\pm$ 0.0013 \\
    \cmidrule(r){2-2}
    \cmidrule(r){3-7}
    \cmidrule(r){8-12}
    & \multirow{2}{*}{FTU-II} & 0.8911 & 0.0240 & 0.0320 & 0.9321 & 1.0* & 0.9920 & 0.0133 & 0.0213 & 0.9735 & 1.0* \\
    & & $\pm$ 0.0045 & $\pm$ 0.0115 & $\pm$ 0.0080 & $\pm$ 0.0229 & $\pm$ 0.0000 & $\pm$ 0.0007 & $\pm$ 0.0013 & $\pm$ 0.0026 & $\pm$ 0.0026 & $\pm$ 0.0000 \\
    \cmidrule(r){2-2}
    \cmidrule(r){3-7}
    \cmidrule(r){8-12}
    & \multirow{2}{*}{CLP} & 0.8911 & 0.0160 & 0.0180 & 0.9682 & 0.9893 & 0.9926 & 0.0120 & 0.0187 & 0.9761 & 0.9974 \\
    & & $\pm$ 0.0133 & $\pm$ 0.0050 & $\pm$ 0.0070 & $\pm$ 0.0099 & $\pm$ 0.0075 & $\pm$ 0.0007 & $\pm$ 0.0013 & $\pm$ 0.0054 & $\pm$ 0.0026 & $\pm$ 0.0014 \\
    \cmidrule(r){2-2}
    \cmidrule(r){3-7}
    \cmidrule(r){8-12}
    & \multirow{2}{*}{$\sysname^\dag$} & 0.9289 & 0.0230 & 0.0220 & 0.9549 & 0.8982 & 0.9953 & 0.0093 & 0.0133 & 0.9814 & 0.9975 \\
    & & $\pm$ 0.0037 & $\pm$ 0.0043 & $\pm$ 0.0053 & $\pm$ 0.0082 & $\pm$ 0.0028 & $\pm$ 0.0010 & $\pm$ 0.0034 & $\pm$ 0.0053 & $\pm$ 0.0065 & $\pm$ 0.0007 \\
    \cmidrule(r){2-2}
    \cmidrule(r){3-7}
    \cmidrule(r){8-12}
    & \multirow{2}{*}{\sysname} & 0.9067 & \textbf{0.0090} & \textbf{0.0080} & \textbf{0.9821} & \textbf{0.9979} & \textbf{0.9953} & \textbf{0.0093} & \textbf{0.0106} & \textbf{0.9815} & \textbf{0.9975} \\
    & & $\pm$ 0.0084 & $\pm$ 0.0010 & $\pm$ 0.0010 & $\pm$ 0.0020 & $\pm$ 0.0006 & $\pm$ 0.0003 & $\pm$ 0.0017 & $\pm$ 0.0027 & $\pm$ 0.0011 & $\pm$ 0.0006 \\
    
    \cmidrule{1-12}
    \multirow{12}{*}{RoBERTa}& \multirow{2}{*}{vanilla} & 0.9200 & 0.1140 & 0.1080 & 0.7989 & 0.8110 & 0.9933  & 0.0106 & 0.0160 & 0.9788 & 0.9924 \\
    & & $\pm$ 0.0055 & $\pm$ 0.0043  & $\pm$ 0.0060 & $\pm$ 0.0112 & $\pm$ 0.0042 & $\pm$ 0.0020 & $\pm$ 0.0027 & $\pm$ 0.0027 & $\pm$ 0.0043 & $\pm$ 0.0016 \\
    \cmidrule(r){2-2}
    \cmidrule(r){3-7}
    \cmidrule(r){8-12}
    & \multirow{2}{*}{FTU-I} & 0.8955 & 0.0869 & 0.0260 & 0.8489 & 0.8964 & 0.9920 & 0.0133 & 0.0160 & 0.9737 & 0.9962 \\
    & & $\pm$ 0.0105 & $\pm$ 0.0275 & $\pm$ 0.0120 & $\pm$ 0.0318 & $\pm$ 0.0298 & $\pm$ 0.0040 & $\pm$ 0.0026 & $\pm$ 0.0027 & $\pm$ 0.0053 & $\pm$ 0.0013 \\
    \cmidrule(r){2-2}
    \cmidrule(r){3-7}
    \cmidrule(r){8-12}
    & \multirow{2}{*}{FTU-II} & 0.8944 & 0.0370 & 0.0260 & 0.9296 & 1.0* & 0.9913 & 0.0093 & 0.0106 & 0.9815 & 1.0* \\
    & & $\pm$ 0.0077 & $\pm$ 0.0220 & $\pm$ 0.0140 & $\pm$ 0.0406 & $\pm$ 0.0000 & $\pm$ 0.0034 & $\pm$ 0.0027 & $\pm$ 0.0040 & $\pm$ 0.0051 & $\pm$ 0.0000 \\
    \cmidrule(r){2-2}
    \cmidrule(r){3-7}
    \cmidrule(r){8-12}
    & \multirow{2}{*}{CLP} & 0.9022 & 0.0160 & 0.0140 & 0.9685 & 0.9888 & 0.9940 & 0.0093 & 0.0160 & 0.9814 & \textbf{0.9987} \\
    & & $\pm$ 0.0055 & $\pm$ 0.0140 & $\pm$ 0.0119 & $\pm$ 0.0258 & $\pm$ 0.0048 & $\pm$ 0.0008  & $\pm$ 0.0020  & $\pm$ 0.0010 & $\pm$ 0.0039 & $\pm$ 0.0013 \\
    \cmidrule(r){2-2}
    \cmidrule(r){3-7}
    \cmidrule(r){8-12}
    & \multirow{2}{*}{$\sysname^\dag$} & \textbf{0.9266} & 0.0240 & 0.0160 & 0.9533 & 0.8826 & 0.9940 & 0.0067 & \textbf{0.0053} & 0.9868 & 0.9950 \\
    & & $\pm$ 0.0033 & $\pm$ 0.0030 & $\pm$ 0.0110 & $\pm$ 0.0060 & $\pm$ 0.0061 & $\pm$ 0.0007 & $\pm$ 0.0012 & $\pm$ 0.0036 & $\pm$ 0.0027 & $\pm$ 0.0031 \\
    \cmidrule(r){2-2}
    \cmidrule(r){3-7}
    \cmidrule(r){8-12}
    & \multirow{2}{*}{\sysname} & 0.9089 & \textbf{0.0060} & \textbf{0.0060} & \textbf{0.9881} & \textbf{0.9973} & \textbf{0.9967} & \textbf{0.0040} & 0.0080 & \textbf{0.9920} & 0.9975 \\
    & & $\pm$ 0.0082 & $\pm$ 0.0020 & $\pm$ 0.0027 & $\pm$ 0.0040 & $\pm$ 0.0015 & $\pm$ 0.0011 & $\pm$ 0.0013 & $\pm$ 0.0022 & $\pm$ 0.0013 & $\pm$ 0.0012 \\
    \bottomrule[1.25pt]
    \end{tabular}
    }
\end{table*}

\subsection{Quantitative Evaluation of Fairness}
\textbf{Fairness and Utility.} We then quantitatively evaluate the effectiveness of \sysname and the compared baselines in terms of model utility and fairness.
The main evaluation results are summarized in Table.\ref{tab:main_result}.
Notice that the values of these five metrics range from 0 to 1. 
A higher value of Acc. reflects a better model utility. A smaller value of DPD and EOD, and a higher value of DIR and CTF indicate a better fairness.
All evaluation experiments are conducted on the whole testing datasets.

It can be obviously seen from Table.\ref{tab:main_result} that all vanilla models have both achieved considerable high accuracy in the target classification tasks (i.e., the accuracy is above 0.9200 on toxicity detection and above 0.9900 on sentiment analysis), indicating good model utility.
It also suggests that the unfairness issue in models trained on Jigsaw Toxicity is more severe than those on Yelp Sentiment.
We speculate that this is mainly because people are often mixed with more social prejudices and personal stereotypes when expressing insulting comments (e.g., hate speech and racial discrimination), resulting in more unfair issues in the toxicity detection tasks.
After applying FTU-I as the mitigation method, a slight increase in fairness is gained at the sacrifice of model accuracy.
For instance, the accuracy of vanilla RoBERTa on Jigsaw Toxicity dropped from $0.9200$ to $0.8955$ with a subtle improvement of $0.0271$ in DPD.
When FTU-II is used for unfairness mitigation, the promotion in fairness grows further but along with a larger drop in model accuracy, which may result in a trade-off between model utility and fairness.
Similarly, a trade-off between model accuracy and fairness is also observed when employing the CLP mitigation method although it obtains a relatively significant improvement in terms of the four fairness dimensions.  
We speculate this is mainly because that CLP is also subject to the drawbacks of missing consideration in correlations between sensitive attributes and non-sensitive attributes, and the added robust term in optimization objective has limited the model's ability in capturing the semantics of input.

\begin{figure*}[tp]
\centering
\subfigure[Jigsaw Toxicity]{
    \centering
    \includegraphics[width=0.45\textwidth]{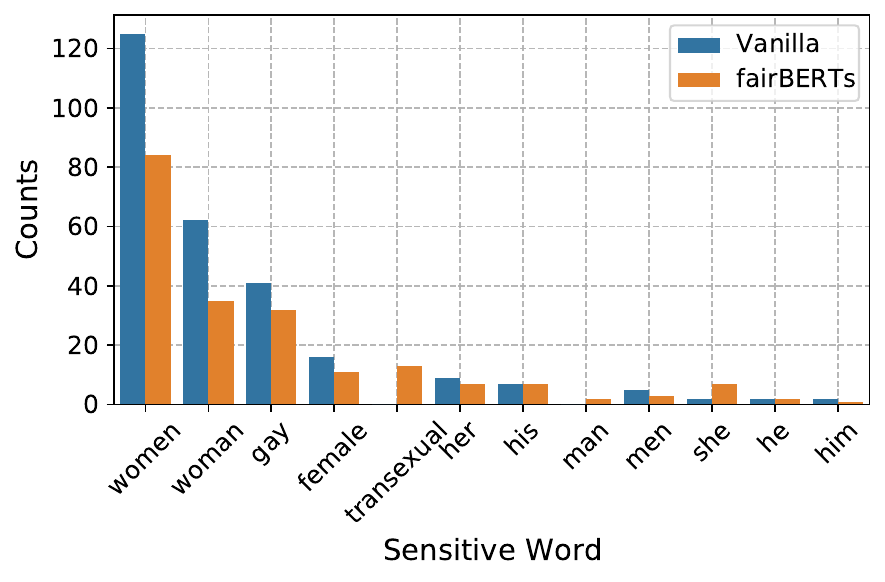}
    \label{fig:sensitivity_drop_jigsaw}
}
\subfigure[Yelp Sentiment]{
    \centering
    \includegraphics[width=0.45\textwidth]{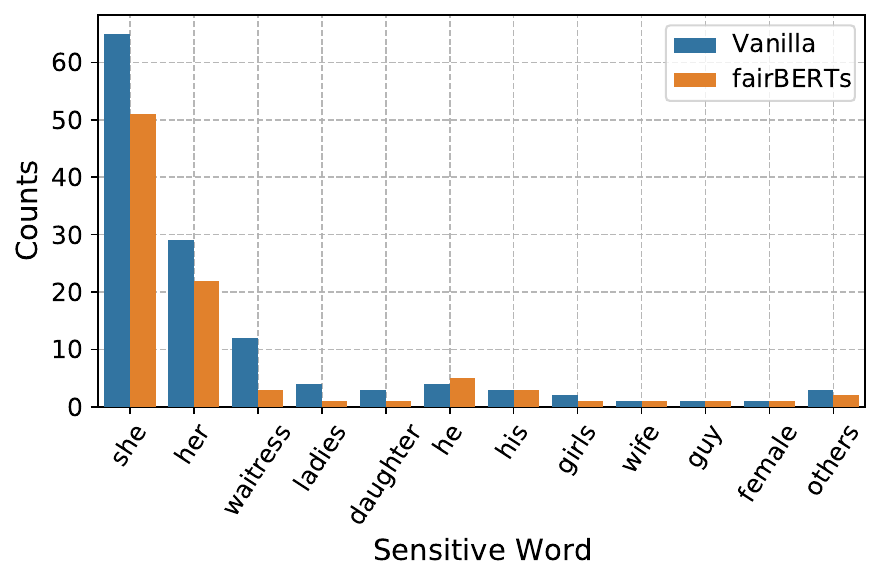}
    \label{fig:sensitivity_drop_yelp}
}
\caption{Comparison of sensitive words in the top-3 important decision words given by LIME.}
\label{fig:sensi_words_comparison}
\end{figure*}

In contrast, \sysname learnt without counterfactual adversarial training has achieved fairness gains comparable to the CLP method, and almost has no negative impact on the model utility.
In most cases, the model accuracy has been promoted instead along with the improvement in model fairness.
Specifically, the accuracy of BERT and RoBERTa on Yelp Sentiment has been promoted from 0.9920 and 0.9933 to 0.9953 and 0.9940, respectively.
This is mainly attributed to the generative adversarial framework, which has also been proved in \cite{croce2020gan}.
When learning with counterfactual adversarial training, \sysname has exhibited great superiority in mitigating the involved unfairness and outperforms the compared baselines by a significant margin.
This demonstrates that \sysname is effective in mitigating the bias encoded in model while balancing model utility well in the meantime.

\textbf{Reduction of Sensitive Words.}
We further analyze the changement of sensitive words in the top-3 important decision-making words before and after bias mitigation by taking the BERT backbone as an example.
The quantitative results on the two datasets are shown in Figure.\ref{fig:sensi_words_comparison}, in which the top-3 important decision words are given by LIME through model-agnostic interpretation.

It is clearly observed that there is a significant decrease in the twelve representative sensitive gender words frequently occurred in the top-3 important decision words.
Concretely, the amount of gender words decreases by above $30\%$ on the toxicity detection task and a similar trend is shown on the sentiment analysis task.
This suggests that \sysname can reduce the dependence of model decision on the protected sensitive information, which would help models focus on the real feature in decision-making.

\subsection{Evaluation of Transferability}
Finally, we further investigate the transferability of the generator $G$ as well as the generated semantic and fairness-aware perturbation $\bm{\delta}$.
Specifically, we directly apply the corresponding perturbation $\bm{\delta}$ generated by $G$ to the classification representation of the conventionally trained vanilla models for each input text sample, and then reevaluate the utility and fairness of vanilla models from the five aspects on the testing sets.
We take the toxicity detection task as an example to report the evaluation results and the main experimental results are summarized in Table.\ref{tab:transfer_evaluation}.

Compared to the vanilla BERT and RoBERTa, after applying the semantic and fairness-aware perturbations, the two models both attain a noticeable improvement in the four fairness metrics while maintaining most of the model utility.
Specifically, the DIR of BERT and RoBERTa is promoted from 0.8716 and 0.7989 to 0.9206 and 0.9318 respectively with just a small drop of about 2\% to 3\% in accuracy.
This transferred performance is almost comparable to that of the FTU-II method, which demonstrates that the generated semantic and fairness-aware perturbation is indeed transferable in term of fairness.
We can therefore conclude that it is completely feasible to transfer the adversarial components in \sysname to other models for achieving fairness improvements.

\begin{table}[tp]
    \renewcommand\arraystretch{1.05}
    \setlength\tabcolsep{3pt}
    \caption{Evaluation in transferability of semantic and fairness-aware perturbation generated in \sysname.}
    \label{tab:transfer_evaluation}
    \resizebox{0.47\textwidth}{!}{
    \centering
    \begin{tabular}{lrrrrr}
    \toprule[1.25pt]
    \multirow{2}{*}{\textbf{Model}} & \multicolumn{5}{c}{\textbf{Jigsaw Toxicity}} \\
    \cmidrule(r){2-6}
    & Acc. & DPD & EOD & DIR & CTF  \\
    \cmidrule{1-6}
    \multirow{2}{*}{BERT}& 0.9326 & 0.0670 & 0.0820 & 0.8716 & 0.8481 \\
    & $\pm$ 0.0076 & $\pm$ 0.0062 & $\pm$ 0.0054 & $\pm$ 0.0105 & $\pm$ 0.0036  \\
    \cmidrule(r){1-1}
    \cmidrule(r){2-6}
    \multirow{2}{*}{BERT + $\bm{\delta}$} & 0.9065 & 0.0324  & 0.0345 & 0.9206  & 0.8897  \\
    & $\pm$ 0.0053  & $\pm$ 0.0150  & $\pm$ 0.0102  & $\pm$ 0.0145 & $\pm$ 0.0068 \\
    
    \cmidrule{1-6}
    \multirow{2}{*}{RoBERTa}& 0.9200 & 0.1140 & 0.1080 & 0.7989 & 0.8110  \\
    & $\pm$ 0.0055 & $\pm$ 0.0043  & $\pm$ 0.0060 & $\pm$ 0.0112 & $\pm$ 0.0042  \\
    \cmidrule(r){1-1}
    \cmidrule(r){2-6}
    \multirow{2}{*}{RoBERTa + $\bm{\delta}$} &  0.9047 & 0.0515 & 0.0426  & 0.9318  & 0.8564 \\
    & $\pm$ 0.0061  & $\pm$ 0.0083  & $\pm$ 0.0124  & $\pm$ 0.0196  & $\pm$ 0.0093 \\
    \bottomrule[1.25pt]
    \end{tabular}
    }
\end{table}

\section{Conclusion}
To solve the problem that protected sensitive information leaks into intermediate model representations, we present \sysname, a general framework for building fair BERT series fine-tuned models by erasing sensitive information via semantic and fairness-aware perturbations.
Through qualitative and quantitative empirical evaluation, we demonstrate that \sysname attains promising effectiveness in mitigating the unfairness involved in the model decision-making process while preserves the model utility well at the mean time.
We also show the feasibility of transferring the adversarial debiasing component in \sysname to other conventionally trained models for yielding fairness improvements.
A promising future work is to generalize \sysname to other tasks or research domains for building more trustworthy fine-tuned PLMs.

\newpage
\bibliographystyle{ACM-Reference-Format}
\bibliography{references}


\begin{thebibliography}{35}


\ifx \showCODEN    \undefined \def \showCODEN     #1{\unskip}     \fi
\ifx \showDOI      \undefined \def \showDOI       #1{#1}\fi
\ifx \showISBNx    \undefined \def \showISBNx     #1{\unskip}     \fi
\ifx \showISBNxiii \undefined \def \showISBNxiii  #1{\unskip}     \fi
\ifx \showISSN     \undefined \def \showISSN      #1{\unskip}     \fi
\ifx \showLCCN     \undefined \def \showLCCN      #1{\unskip}     \fi
\ifx \shownote     \undefined \def \shownote      #1{#1}          \fi
\ifx \showarticletitle \undefined \def \showarticletitle #1{#1}   \fi
\ifx \showURL      \undefined \def \showURL       {\relax}        \fi
\providecommand\bibfield[2]{#2}
\providecommand\bibinfo[2]{#2}
\providecommand\natexlab[1]{#1}
\providecommand\showeprint[2][]{arXiv:#2}

\bibitem[Angwin et~al\mbox{.}(2016)]%
        {machine_bias}
\bibfield{author}{\bibinfo{person}{Julia Angwin}, \bibinfo{person}{Jeff
  Larson}, \bibinfo{person}{Surya Mattu}, {and} \bibinfo{person}{Lauren
  Kirchner}.} \bibinfo{year}{2016}\natexlab{}.
\newblock \bibinfo{title}{Machine Bias: There’s software used across the
  country to predict future criminals. And it’s biased against blacks.}
\newblock
\newblock
\urldef\tempurl%
\url{https://www.bloomberg.com/news/articles/2019-06-18/facebook-warns-it-can-t-fully-solve-toxic-content-problem}
\showURL{%
\tempurl}


\bibitem[Beutel et~al\mbox{.}(2017)]%
        {beutel2017data}
\bibfield{author}{\bibinfo{person}{Alex Beutel}, \bibinfo{person}{Jilin Chen},
  \bibinfo{person}{Zhe Zhao}, {and} \bibinfo{person}{Ed~H Chi}.}
  \bibinfo{year}{2017}\natexlab{}.
\newblock \showarticletitle{Data decisions and theoretical implications when
  adversarially learning fair representations}.
\newblock \bibinfo{journal}{\emph{arXiv preprint arXiv:1707.00075}}
  (\bibinfo{year}{2017}).
\newblock


\bibitem[Bolukbasi et~al\mbox{.}(2016)]%
        {bolukbasi2016man}
\bibfield{author}{\bibinfo{person}{Tolga Bolukbasi}, \bibinfo{person}{Kai-Wei
  Chang}, \bibinfo{person}{James~Y Zou}, \bibinfo{person}{Venkatesh Saligrama},
  {and} \bibinfo{person}{Adam~T Kalai}.} \bibinfo{year}{2016}\natexlab{}.
\newblock \showarticletitle{Man is to Computer Programmer as Woman is to
  Homemaker? Debiasing Word Embeddings}. In
  \bibinfo{booktitle}{\emph{Proceedings of the 30th International Conference on
  Neural Information Processing Systemss}}. \bibinfo{pages}{4356--4364}.
\newblock


\bibitem[Caliskan et~al\mbox{.}(2017)]%
        {caliskan2017semantics}
\bibfield{author}{\bibinfo{person}{Aylin Caliskan}, \bibinfo{person}{Joanna
  Bryson}, {and} \bibinfo{person}{Arvind Narayanan}.}
  \bibinfo{year}{2017}\natexlab{}.
\newblock \showarticletitle{Semantics derived automatically from language
  corpora contain human-like biases}.
\newblock \bibinfo{journal}{\emph{Science}}  \bibinfo{volume}{356}
  (\bibinfo{date}{04} \bibinfo{year}{2017}), \bibinfo{pages}{183--186}.
\newblock


\bibitem[Clark et~al\mbox{.}(2020)]%
        {clark2020electra}
\bibfield{author}{\bibinfo{person}{Kevin Clark}, \bibinfo{person}{Minh-Thang
  Luong}, \bibinfo{person}{Quoc~V. Le}, {and} \bibinfo{person}{Christopher~D.
  Manning}.} \bibinfo{year}{2020}\natexlab{}.
\newblock \showarticletitle{{ELECTRA}: Pre-training Text Encoders as
  Discriminators Rather Than Generators}. In \bibinfo{booktitle}{\emph{ICLR}}.
\newblock
\urldef\tempurl%
\url{https://openreview.net/pdf?id=r1xMH1BtvB}
\showURL{%
\tempurl}


\bibitem[Croce et~al\mbox{.}(2020)]%
        {croce2020gan}
\bibfield{author}{\bibinfo{person}{Danilo Croce}, \bibinfo{person}{Giuseppe
  Castellucci}, {and} \bibinfo{person}{Roberto Basili}.}
  \bibinfo{year}{2020}\natexlab{}.
\newblock \showarticletitle{GAN-BERT: Generative adversarial learning for
  robust text classification with a bunch of labeled examples}. In
  \bibinfo{booktitle}{\emph{Proceedings of the 58th annual meeting of the
  association for computational linguistics}}. \bibinfo{pages}{2114--2119}.
\newblock


\bibitem[de~Vassimon~Manela et~al\mbox{.}(2021)]%
        {de2021stereotype}
\bibfield{author}{\bibinfo{person}{Daniel de Vassimon~Manela},
  \bibinfo{person}{David Errington}, \bibinfo{person}{Thomas Fisher},
  \bibinfo{person}{Boris van Breugel}, {and} \bibinfo{person}{Pasquale
  Minervini}.} \bibinfo{year}{2021}\natexlab{}.
\newblock \showarticletitle{Stereotype and skew: Quantifying gender bias in
  pre-trained and fine-tuned language models}. In
  \bibinfo{booktitle}{\emph{EACL 2021-16th Conference of the European Chapter
  of the Association for Computational Linguistics, Proceedings of the
  Conference}}. Association for Computational Linguistics,
  \bibinfo{pages}{2232--2242}.
\newblock


\bibitem[Deng and Tian(2019)]%
        {deng2019towards}
\bibfield{author}{\bibinfo{person}{Congyue Deng} {and} \bibinfo{person}{Yi
  Tian}.} \bibinfo{year}{2019}\natexlab{}.
\newblock \showarticletitle{Towards Understanding the Trade-off Between
  Accuracy and Adversarial Robustness}. In \bibinfo{booktitle}{\emph{ICML 2019
  Workshop on Security and Privacy of Machine Learning}}.
\newblock


\bibitem[Devlin et~al\mbox{.}(2019)]%
        {devlin-etal-2019-bert}
\bibfield{author}{\bibinfo{person}{Jacob Devlin}, \bibinfo{person}{Ming-Wei
  Chang}, \bibinfo{person}{Kenton Lee}, {and} \bibinfo{person}{Kristina
  Toutanova}.} \bibinfo{year}{2019}\natexlab{}.
\newblock \showarticletitle{{BERT}: Pre-training of Deep Bidirectional
  Transformers for Language Understanding}. In
  \bibinfo{booktitle}{\emph{Proceedings of the 2019 Conference of the North
  {A}merican Chapter of the Association for Computational Linguistics: Human
  Language Technologies, Volume 1 (Long and Short Papers)}}.
  \bibinfo{pages}{4171--4186}.
\newblock


\bibitem[Dinan et~al\mbox{.}(2020)]%
        {dinan-etal-2020-multi}
\bibfield{author}{\bibinfo{person}{Emily Dinan}, \bibinfo{person}{Angela Fan},
  \bibinfo{person}{Ledell Wu}, \bibinfo{person}{Jason Weston},
  \bibinfo{person}{Douwe Kiela}, {and} \bibinfo{person}{Adina Williams}.}
  \bibinfo{year}{2020}\natexlab{}.
\newblock \showarticletitle{Multi-Dimensional Gender Bias Classification}. In
  \bibinfo{booktitle}{\emph{Proceedings of the 2020 Conference on Empirical
  Methods in Natural Language Processing (EMNLP)}}.
  \bibinfo{publisher}{Association for Computational Linguistics},
  \bibinfo{address}{Online}, \bibinfo{pages}{314--331}.
\newblock


\bibitem[Dixon et~al\mbox{.}(2018)]%
        {dixon2018measuring}
\bibfield{author}{\bibinfo{person}{Lucas Dixon}, \bibinfo{person}{John Li},
  \bibinfo{person}{Jeffrey Sorensen}, \bibinfo{person}{Nithum Thain}, {and}
  \bibinfo{person}{Lucy Vasserman}.} \bibinfo{year}{2018}\natexlab{}.
\newblock \showarticletitle{Measuring and mitigating unintended bias in text
  classification}. In \bibinfo{booktitle}{\emph{Proceedings of the 2018
  AAAI/ACM Conference on AI, Ethics, and Society}}. \bibinfo{pages}{67--73}.
\newblock


\bibitem[Dwork et~al\mbox{.}(2012)]%
        {dwork2012fairness}
\bibfield{author}{\bibinfo{person}{Cynthia Dwork}, \bibinfo{person}{Moritz
  Hardt}, \bibinfo{person}{Toniann Pitassi}, \bibinfo{person}{Omer Reingold},
  {and} \bibinfo{person}{Richard Zemel}.} \bibinfo{year}{2012}\natexlab{}.
\newblock \showarticletitle{Fairness through awareness}. In
  \bibinfo{booktitle}{\emph{Proceedings of the 3rd innovations in theoretical
  computer science conference}}. \bibinfo{pages}{214--226}.
\newblock


\bibitem[Elazar and Goldberg(2018)]%
        {elazar2018adversarial}
\bibfield{author}{\bibinfo{person}{Yanai Elazar} {and} \bibinfo{person}{Yoav
  Goldberg}.} \bibinfo{year}{2018}\natexlab{}.
\newblock \showarticletitle{Adversarial removal of demographic attributes from
  text data}.
\newblock \bibinfo{journal}{\emph{arXiv preprint arXiv:1808.06640}}
  (\bibinfo{year}{2018}).
\newblock


\bibitem[Feldman et~al\mbox{.}(2015)]%
        {feldman2015certifying}
\bibfield{author}{\bibinfo{person}{Michael Feldman}, \bibinfo{person}{Sorelle~A
  Friedler}, \bibinfo{person}{John Moeller}, \bibinfo{person}{Carlos
  Scheidegger}, {and} \bibinfo{person}{Suresh Venkatasubramanian}.}
  \bibinfo{year}{2015}\natexlab{}.
\newblock \showarticletitle{Certifying and removing disparate impact}. In
  \bibinfo{booktitle}{\emph{proceedings of the 21th ACM SIGKDD international
  conference on knowledge discovery and data mining}}.
  \bibinfo{pages}{259--268}.
\newblock


\bibitem[Gajane and Pechenizkiy(2017)]%
        {gajane2017formalizing}
\bibfield{author}{\bibinfo{person}{Pratik Gajane} {and} \bibinfo{person}{Mykola
  Pechenizkiy}.} \bibinfo{year}{2017}\natexlab{}.
\newblock \showarticletitle{On formalizing fairness in prediction with machine
  learning}.
\newblock \bibinfo{journal}{\emph{arXiv preprint arXiv:1710.03184}}
  (\bibinfo{year}{2017}).
\newblock


\bibitem[Garg et~al\mbox{.}(2019)]%
        {garg2019counterfactual}
\bibfield{author}{\bibinfo{person}{Sahaj Garg}, \bibinfo{person}{Vincent
  Perot}, \bibinfo{person}{Nicole Limtiaco}, \bibinfo{person}{Ankur Taly},
  \bibinfo{person}{Ed~H Chi}, {and} \bibinfo{person}{Alex Beutel}.}
  \bibinfo{year}{2019}\natexlab{}.
\newblock \showarticletitle{Counterfactual fairness in text classification
  through robustness}. In \bibinfo{booktitle}{\emph{Proceedings of the 2019
  AAAI/ACM Conference on AI, Ethics, and Society}}. \bibinfo{pages}{219--226}.
\newblock


\bibitem[Goodfellow et~al\mbox{.}(2020)]%
        {goodfellow2020generative}
\bibfield{author}{\bibinfo{person}{Ian Goodfellow}, \bibinfo{person}{Jean
  Pouget-Abadie}, \bibinfo{person}{Mehdi Mirza}, \bibinfo{person}{Bing Xu},
  \bibinfo{person}{David Warde-Farley}, \bibinfo{person}{Sherjil Ozair},
  \bibinfo{person}{Aaron Courville}, {and} \bibinfo{person}{Yoshua Bengio}.}
  \bibinfo{year}{2020}\natexlab{}.
\newblock \showarticletitle{Generative adversarial networks}.
\newblock \bibinfo{journal}{\emph{Commun. ACM}} \bibinfo{volume}{63},
  \bibinfo{number}{11} (\bibinfo{year}{2020}), \bibinfo{pages}{139--144}.
\newblock


\bibitem[Goodfellow et~al\mbox{.}(2015)]%
        {goodfellow2014explaining}
\bibfield{author}{\bibinfo{person}{Ian~J Goodfellow}, \bibinfo{person}{Jonathon
  Shlens}, {and} \bibinfo{person}{Christian Szegedy}.}
  \bibinfo{year}{2015}\natexlab{}.
\newblock \showarticletitle{Explaining and harnessing adversarial examples}. In
  \bibinfo{booktitle}{\emph{ICLR}}.
\newblock


\bibitem[Hardt et~al\mbox{.}(2016)]%
        {hardt2016equality}
\bibfield{author}{\bibinfo{person}{Moritz Hardt}, \bibinfo{person}{Eric Price},
  {and} \bibinfo{person}{Nathan Srebro}.} \bibinfo{year}{2016}\natexlab{}.
\newblock \showarticletitle{Equality of opportunity in supervised learning}. In
  \bibinfo{booktitle}{\emph{Proceedings of the 30th International Conference on
  Neural Information Processing Systems}}. \bibinfo{pages}{3323--3331}.
\newblock


\bibitem[Hutchinson et~al\mbox{.}(2020)]%
        {hutchinson2020social}
\bibfield{author}{\bibinfo{person}{Ben Hutchinson}, \bibinfo{person}{Vinodkumar
  Prabhakaran}, \bibinfo{person}{Emily Denton}, \bibinfo{person}{Kellie
  Webster}, \bibinfo{person}{Yu Zhong}, {and} \bibinfo{person}{Stephen
  Denuyl}.} \bibinfo{year}{2020}\natexlab{}.
\newblock \showarticletitle{Social Biases in NLP Models as Barriers for Persons
  with Disabilities}. In \bibinfo{booktitle}{\emph{Proceedings of the 58th
  Annual Meeting of the Association for Computational Linguistics}}.
  \bibinfo{pages}{5491--5501}.
\newblock


\bibitem[Kusner et~al\mbox{.}(2017)]%
        {kusner2017counterfactual}
\bibfield{author}{\bibinfo{person}{Matt Kusner}, \bibinfo{person}{Joshua
  Loftus}, \bibinfo{person}{Chris Russell}, {and} \bibinfo{person}{Ricardo
  Silva}.} \bibinfo{year}{2017}\natexlab{}.
\newblock \showarticletitle{Counterfactual fairness}. In
  \bibinfo{booktitle}{\emph{Proceedings of the 31st International Conference on
  Neural Information Processing Systems}}. \bibinfo{pages}{4069--4079}.
\newblock


\bibitem[Liang et~al\mbox{.}(2020)]%
        {liang2020monolingual}
\bibfield{author}{\bibinfo{person}{Sheng Liang}, \bibinfo{person}{Philipp
  Dufter}, {and} \bibinfo{person}{Hinrich Sch{\"u}tze}.}
  \bibinfo{year}{2020}\natexlab{}.
\newblock \showarticletitle{Monolingual and Multilingual Reduction of Gender
  Bias in Contextualized Representations}. In
  \bibinfo{booktitle}{\emph{Proceedings of the 28th International Conference on
  Computational Linguistics}}. \bibinfo{pages}{5082--5093}.
\newblock


\bibitem[Liu et~al\mbox{.}(2019)]%
        {liu2019roberta}
\bibfield{author}{\bibinfo{person}{Yinhan Liu}, \bibinfo{person}{Myle Ott},
  \bibinfo{person}{Naman Goyal}, \bibinfo{person}{Jingfei Du},
  \bibinfo{person}{Mandar Joshi}, \bibinfo{person}{Danqi Chen},
  \bibinfo{person}{Omer Levy}, \bibinfo{person}{Mike Lewis},
  \bibinfo{person}{Luke Zettlemoyer}, {and} \bibinfo{person}{Veselin
  Stoyanov}.} \bibinfo{year}{2019}\natexlab{}.
\newblock \showarticletitle{Roberta: A robustly optimized bert pretraining
  approach}.
\newblock \bibinfo{journal}{\emph{arXiv preprint arXiv:1907.11692}}
  (\bibinfo{year}{2019}).
\newblock


\bibitem[Maas et~al\mbox{.}(2013)]%
        {maas2013rectifier}
\bibfield{author}{\bibinfo{person}{Andrew~L Maas}, \bibinfo{person}{Awni~Y
  Hannun}, \bibinfo{person}{Andrew~Y Ng}, {et~al\mbox{.}}}
  \bibinfo{year}{2013}\natexlab{}.
\newblock \showarticletitle{Rectifier nonlinearities improve neural network
  acoustic models}. In \bibinfo{booktitle}{\emph{ICML}}.
\newblock


\bibitem[Park et~al\mbox{.}(2018)]%
        {parketal2018reducing}
\bibfield{author}{\bibinfo{person}{Ji~Ho Park}, \bibinfo{person}{Jamin Shin},
  {and} \bibinfo{person}{Pascale Fung}.} \bibinfo{year}{2018}\natexlab{}.
\newblock \showarticletitle{Reducing Gender Bias in Abusive Language
  Detection}. In \bibinfo{booktitle}{\emph{Proceedings of the 2018 Conference
  on Empirical Methods in Natural Language Processing}}.
  \bibinfo{pages}{2799--2804}.
\newblock


\bibitem[Qian et~al\mbox{.}(2022)]%
        {qian2022perturbation}
\bibfield{author}{\bibinfo{person}{Rebecca Qian}, \bibinfo{person}{Candace
  Ross}, \bibinfo{person}{Jude Fernandes}, \bibinfo{person}{Eric Smith},
  \bibinfo{person}{Douwe Kiela}, {and} \bibinfo{person}{Adina Williams}.}
  \bibinfo{year}{2022}\natexlab{}.
\newblock \showarticletitle{Perturbation augmentation for fairer nlp}.
\newblock \bibinfo{journal}{\emph{arXiv preprint arXiv:2205.12586}}
  (\bibinfo{year}{2022}).
\newblock


\bibitem[Raghunathan et~al\mbox{.}(2020)]%
        {raghunathan2020understanding}
\bibfield{author}{\bibinfo{person}{Aditi Raghunathan},
  \bibinfo{person}{Sang~Michael Xie}, \bibinfo{person}{Fanny Yang},
  \bibinfo{person}{John Duchi}, {and} \bibinfo{person}{Percy Liang}.}
  \bibinfo{year}{2020}\natexlab{}.
\newblock \showarticletitle{Understanding and mitigating the tradeoff between
  robustness and accuracy}. In \bibinfo{booktitle}{\emph{ICML}}.
\newblock


\bibitem[Reddy et~al\mbox{.}(2021)]%
        {reddy2021benchmarking}
\bibfield{author}{\bibinfo{person}{Charan Reddy}, \bibinfo{person}{Deepak
  Sharma}, \bibinfo{person}{Soroush Mehri}, \bibinfo{person}{Adriana
  Romero-Soriano}, \bibinfo{person}{Samira Shabanian}, {and}
  \bibinfo{person}{Sina Honari}.} \bibinfo{year}{2021}\natexlab{}.
\newblock \showarticletitle{Benchmarking bias mitigation algorithms in
  representation learning through fairness metrics}. In
  \bibinfo{booktitle}{\emph{Thirty-fifth Conference on Neural Information
  Processing Systems Datasets and Benchmarks Track (Round 1)}}.
\newblock


\bibitem[Ribeiro et~al\mbox{.}(2016)]%
        {ribeiro2016should}
\bibfield{author}{\bibinfo{person}{Marco~Tulio Ribeiro},
  \bibinfo{person}{Sameer Singh}, {and} \bibinfo{person}{Carlos Guestrin}.}
  \bibinfo{year}{2016}\natexlab{}.
\newblock \showarticletitle{" Why should i trust you?" Explaining the
  predictions of any classifier}. In \bibinfo{booktitle}{\emph{Proceedings of
  the 22nd ACM SIGKDD international conference on knowledge discovery and data
  mining}}. \bibinfo{pages}{1135--1144}.
\newblock


\bibitem[Soares et~al\mbox{.}(2022)]%
        {soares2022your}
\bibfield{author}{\bibinfo{person}{Ioana~Baldini Soares},
  \bibinfo{person}{Dennis Wei}, \bibinfo{person}{Karthikeyan~Natesan
  Ramamurthy}, \bibinfo{person}{Moninder Singh}, {and} \bibinfo{person}{Mikhail
  Yurochkin}.} \bibinfo{year}{2022}\natexlab{}.
\newblock \showarticletitle{Your Fairness May Vary: Pretrained Language Model
  Fairness in Toxic Text Classification}. In \bibinfo{booktitle}{\emph{Annual
  Meeting of the Association for Computational Linguistics}}.
\newblock


\bibitem[Subramanian et~al\mbox{.}(2021)]%
        {subramanian2021fairness}
\bibfield{author}{\bibinfo{person}{Shivashankar Subramanian},
  \bibinfo{person}{Afshin Rahimi}, \bibinfo{person}{Timothy Baldwin},
  \bibinfo{person}{Trevor Cohn}, {and} \bibinfo{person}{Lea Frermann}.}
  \bibinfo{year}{2021}\natexlab{}.
\newblock \showarticletitle{Fairness-aware class imbalanced learning}.
\newblock \bibinfo{journal}{\emph{arXiv preprint arXiv:2109.10444}}
  (\bibinfo{year}{2021}).
\newblock


\bibitem[Szegedy et~al\mbox{.}(2014)]%
        {szegedy2013intriguing}
\bibfield{author}{\bibinfo{person}{Christian Szegedy},
  \bibinfo{person}{Wojciech Zaremba}, \bibinfo{person}{Ilya Sutskever},
  \bibinfo{person}{Joan Bruna}, \bibinfo{person}{Dumitru Erhan},
  \bibinfo{person}{Ian Goodfellow}, {and} \bibinfo{person}{Rob Fergus}.}
  \bibinfo{year}{2014}\natexlab{}.
\newblock \showarticletitle{Intriguing properties of neural networks}. In
  \bibinfo{booktitle}{\emph{ICLR}}.
\newblock


\bibitem[Wang et~al\mbox{.}(2022)]%
        {wang2022fairness}
\bibfield{author}{\bibinfo{person}{Zhibo Wang}, \bibinfo{person}{Xiaowei Dong},
  \bibinfo{person}{Henry Xue}, \bibinfo{person}{Zhifei Zhang},
  \bibinfo{person}{Weifeng Chiu}, \bibinfo{person}{Tao Wei}, {and}
  \bibinfo{person}{Kui Ren}.} \bibinfo{year}{2022}\natexlab{}.
\newblock \showarticletitle{Fairness-aware Adversarial Perturbation Towards
  Bias Mitigation for Deployed Deep Models}. In
  \bibinfo{booktitle}{\emph{Proceedings of the IEEE/CVF Conference on Computer
  Vision and Pattern Recognition}}. \bibinfo{pages}{10379--10388}.
\newblock


\bibitem[Wick et~al\mbox{.}(2019)]%
        {wick2019unlocking}
\bibfield{author}{\bibinfo{person}{Michael Wick}, \bibinfo{person}{Swetasudha
  Panda}, {and} \bibinfo{person}{Jean-Baptiste Tristan}.}
  \bibinfo{year}{2019}\natexlab{}.
\newblock \showarticletitle{Unlocking fairness: a trade-off revisited}. In
  \bibinfo{booktitle}{\emph{Proceedings of the 33rd International Conference on
  Neural Information Processing Systems}}. \bibinfo{pages}{8783--8792}.
\newblock


\bibitem[Zhang et~al\mbox{.}(2018)]%
        {zhang2018mitigating}
\bibfield{author}{\bibinfo{person}{Brian~Hu Zhang}, \bibinfo{person}{Blake
  Lemoine}, {and} \bibinfo{person}{Margaret Mitchell}.}
  \bibinfo{year}{2018}\natexlab{}.
\newblock \showarticletitle{Mitigating unwanted biases with adversarial
  learning}. In \bibinfo{booktitle}{\emph{Proceedings of the 2018 AAAI/ACM
  Conference on AI, Ethics, and Society}}. \bibinfo{pages}{335--340}.
\newblock


\end{thebibliography}

\appendix

\end{document}